\documentclass[conference]{IEEEtran}
\IEEEoverridecommandlockouts
\usepackage{cite}
\usepackage[inline]{enumitem}
\usepackage{subfig}
\usepackage{array}
\usepackage{multirow}
\usepackage{algorithm}
\usepackage{algorithmic}
\usepackage{amsmath,amssymb,amsfonts}
\usepackage{graphicx}
\usepackage{textcomp}
\usepackage{float}
\usepackage{booktabs}
\usepackage{subcaption}

\usepackage{color}
\ifCLASSINFOpdf
\else
\fi
\begin{document}

\title{VaCDA: Variational Contrastive Alignment-based Scalable Human Activity Recognition}

\author{
\IEEEauthorblockN{Soham Khisa}
\IEEEauthorblockA{\textit{Department of Computer Science and Engineering} \\
\textit{Bangladesh University of Engineering and Technology}\\
Dhaka, Bangladesh \\
Email: 1705120@ugrad.cse.buet.ac.bd}
\and
\IEEEauthorblockN{Avijoy Chakma}
\IEEEauthorblockA{\textit{Department of Computer Science} \\
\textit{Bowie State University}\\
Maryland, USA \\
Email: achakma@bowiestate.edu}
}

\maketitle
\begin{abstract}
    Technological advancements have led to the rise of wearable devices with sensors that continuously monitor user activities, generating vast amounts of unlabeled data. This data is challenging to interpret, and manual annotation is labor-intensive and error-prone. Additionally, data distribution is often heterogeneous due to device placement, type, and user behavior variations. As a result, traditional transfer learning methods perform suboptimally, making it difficult to recognize daily activities. To address these challenges, we use a variational autoencoder (VAE) to learn a shared, low-dimensional latent space from available sensor data. This space generalizes data across diverse sensors, mitigating heterogeneity and aiding robust adaptation to the target domain. We integrate contrastive learning to enhance feature representation by aligning instances of the same class across domains while separating different classes. We propose Variational Contrastive Domain Adaptation (VaCDA), a multi-source domain adaptation framework combining VAEs and contrastive learning to improve feature representation and reduce heterogeneity between source and target domains. We evaluate VaCDA on multiple publicly available datasets across three heterogeneity scenarios: cross-person, cross-position, and cross-device. VaCDA outperforms the baselines in cross-position and cross-device scenarios.

\end{abstract}

\begin{figure}[t]
\centering
\includegraphics[width=.5\textwidth]{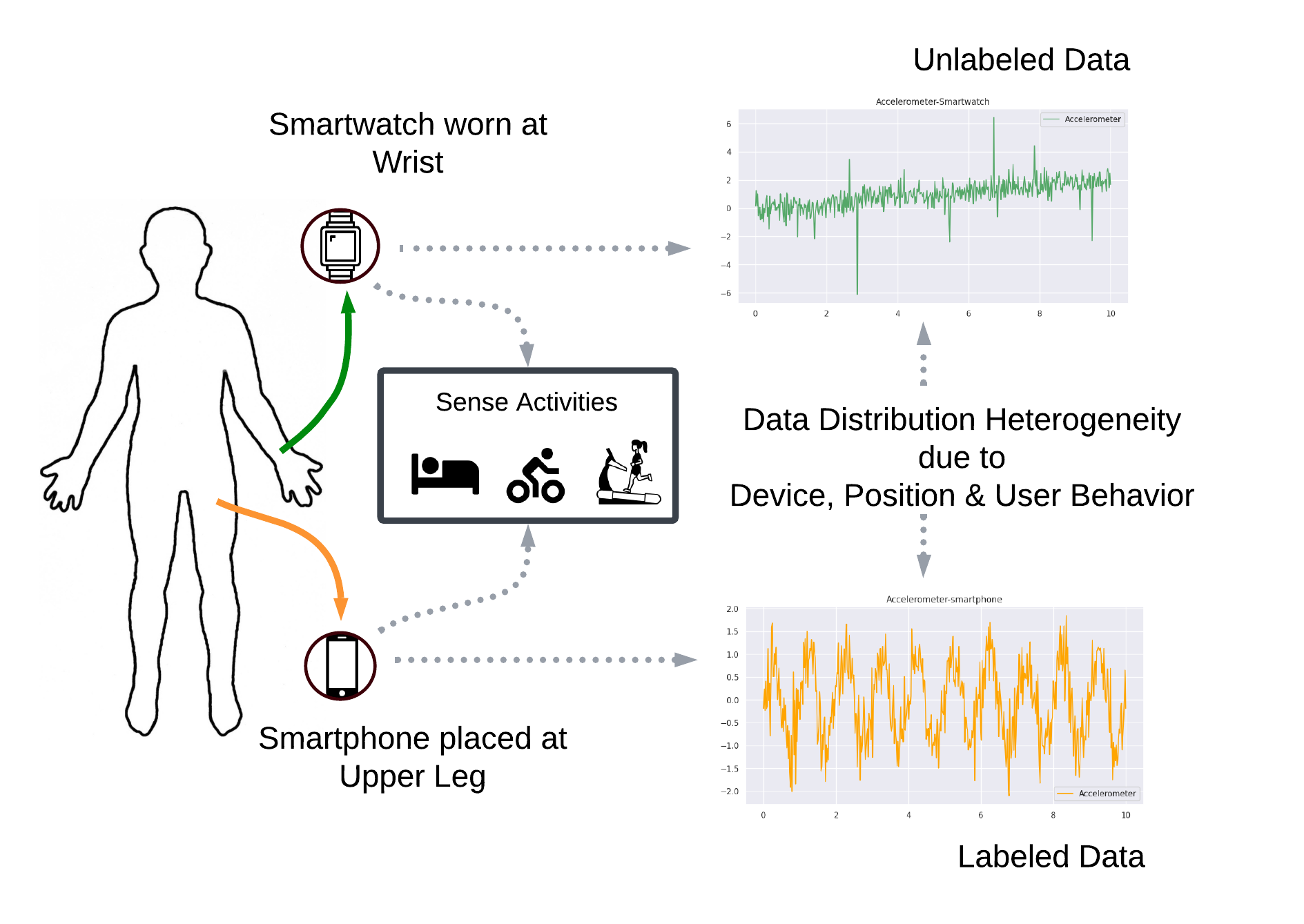}
\caption{An illustration of domain adaptation in human activity recognition. The image highlights data distribution heterogeneity due to device placement and user behavior.}
\label{fig:arc0}
\end{figure}

\section{Introduction}




The proliferation of wearable devices has enabled continuous sensing, processing, and recognition of human activities (HAR). Wearable sensing-based HAR has garnered a wide range of applications such as healthcare~\cite{9055403}, mental health monitoring~\cite{nouman2021recent}, smart homes~\cite{article}, smart textile~\cite{shi2020smart}, sports analytics \cite{HMSajjadHossain2017SoccerMateAP} and so on due to its less privacy intrusive, often user-friendly nature. Along with the boon of sensing capabilities, there exist critical research challenges. Device-integrated sensors continuously sense the user's activities and nearby surroundings, and the sensed data yields a large volume of unlabeled data. Here, the traditional supervised machine learning approach requires labeled data samples for model training, and the same is true for fine-tuning a pre-trained deep learning model. Acquiring label data is often challenging as manual data annotation is cumbersome and error-prone. Note that, machine learning models assume the training and the test dataset follow the same underlying distribution, which is often unrealistic. Thus, due to the data distribution heterogeneity between the training and test data, a machine learning model trained on a labeled dataset performs sub-optimal when evaluated on a relevant but different unlabeled dataset. This phenomenon of data distribution heterogeneity is known as ``domain shift'' or ``distribution shift''~\cite{liu2022deep,chang2020systematic}.



Data distributions can vary due to factors like device positioning, user behavior, and environmental differences. Figure~\ref{fig:arc0} shows wearable data from two devices (smartphone and smartwatch) placed on the dominant wrist and upper leg. The data distribution likely differs because of the varying movement ranges of these body parts, and because different devices use different sensing principles. Additionally, personal factors like age, BMI, and health can affect user behavior, contributing to distribution differences---such as one person’s jogging pattern resembling another’s walking. Thus, incorporating the right modeling components is essential to handle data distribution shifts and build a robust, scalable HAR system.

Domain adaptation (DA) is a branch of transfer learning technique that addresses the data distribution shift~\cite{csurka2017domain}. We consider the ``source domain'' as the labeled dataset(s) and the unlabeled dataset as the ``target domain''~\cite{wilson2020survey}. Aligning the two related data distributions from the source and target domain close to each other is a common approach to reducing the data distribution heterogeneity. However, the problem becomes more challenging with multiple source domains. In real-world scenarios, we often have more than one source, and it’s not always clear which one to use for adaptation. This highlights the need for multi-source domain adaptation (MSDA). Most DA approaches can be grouped under the discriminative approach to address the issue that often incorporates adversarial learning~\cite{8444572,8444585,CHAKMA2021100174,wilson2020multisourcedeepdomainadaptation}, divergence minimization-based techniques~\cite{8444585,CHEN20191} and contrastive learning~\cite{SANABRIA2021101477,wilson2023caldaimprovingmultisourcetime}. 
Contrastive learning differentiates between similar and dissimilar data points without labeled data, pulling positive pairs (e.g., augmented versions of the same sample) closer and pushing negative pairs (e.g., different samples) apart in the learned feature space. While most methods focus on discriminative approaches, fewer explore generative ones. Variational autoencoders (VAE) \cite{kingma2022autoencodingvariationalbayes}, a prominent generative approach, have been shown to effectively capture low-level features across data distributions \cite{pmlr-v139-havtorn21a}. This ability to generalize distributions makes VAE suitable for MSDA. However, VAEs can over-generalize, mapping different activities with similar features to the same space. To address this, we consider contrastive learning.

In this paper, we have made the following contributions:
\begin{enumerate}
\item We propose VaCDA, a novel probabilistic approach that utilizes the feature generalization capability of VAE across domains and integrates the discriminative power of contrastive learning.
\item We propose a scalable multi-source domain adaptation solution to address data distribution heterogeneity. By leveraging the feature generalization capability of VAE, VaCDA can handle an arbitrary number of sources. This approach eliminates the need for separate pipelines for each domain.
\item We perform extensive experiments on four publicly available datasets. We evaluate VaCDA in scenarios involving cross-position, cross-person, and cross-device heterogeneity. Our experiments show that VaCDA outperforms the baselines in cross-position and cross-device settings.
\end{enumerate}

The rest of the paper is organized as follows. In section~\ref{sec:related}, we review the current literature on domain adaptation in activity recognition. Section~\ref{sec:vacda} describes the VaCDA framework. Section~\ref{sec:experiments} details the design choices of our experiment, describes the datasets, and discusses the baseline machine learning approaches. In section~\ref{sec:result&analysis}, we discuss the results of the VaCDA framework and finally, we conclude in section~\ref{sec:conclusion}.

\begin{figure*}[t]
\centering
\includegraphics[width=.8\textwidth]{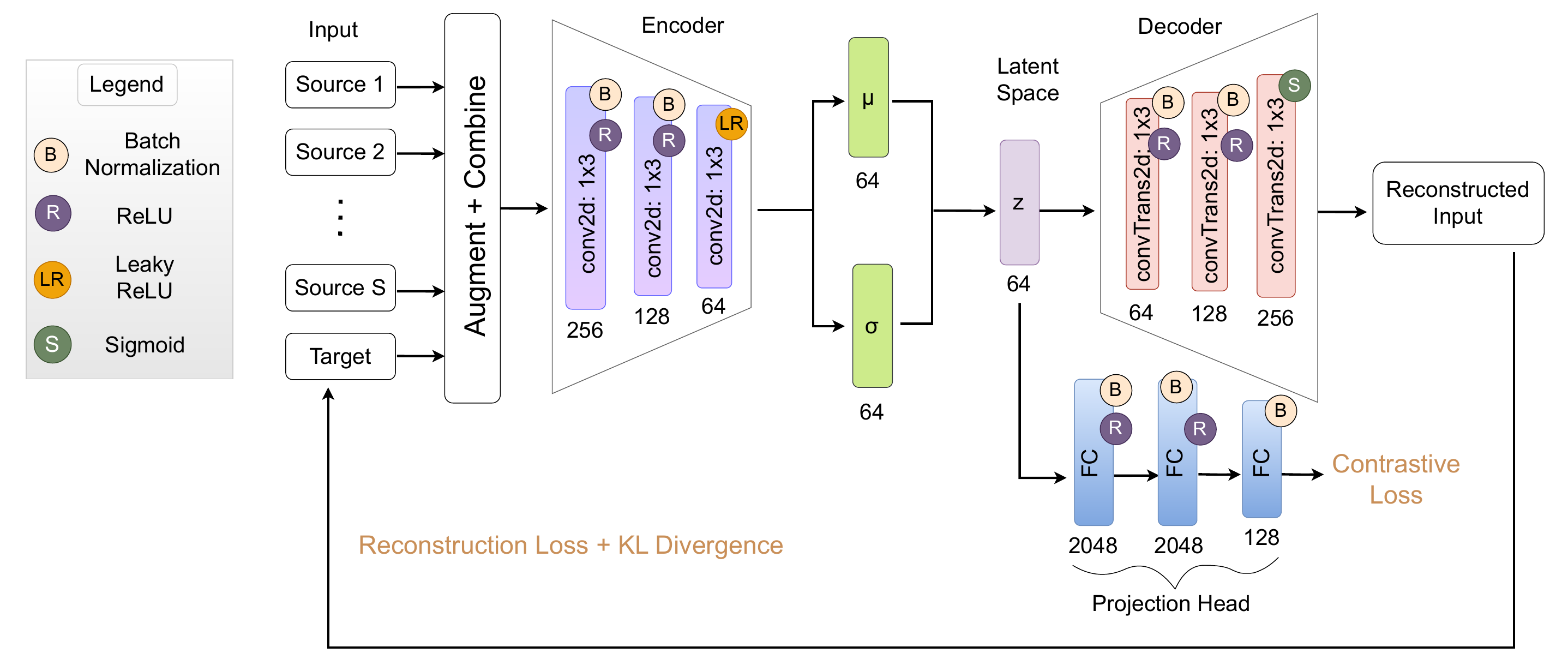}
\vspace{0.5em}
\caption{VaCDA pertaining pipeline; it combines VAE and contrastive learning. The encoder processes available data from all domains. The framework optimizes reconstruction loss, KL divergence, and contrastive loss for domain adaptation.}
\label{fig:arc1}
\end{figure*}

\begin{figure*}[t]
\centering
\includegraphics[width=0.8\textwidth]{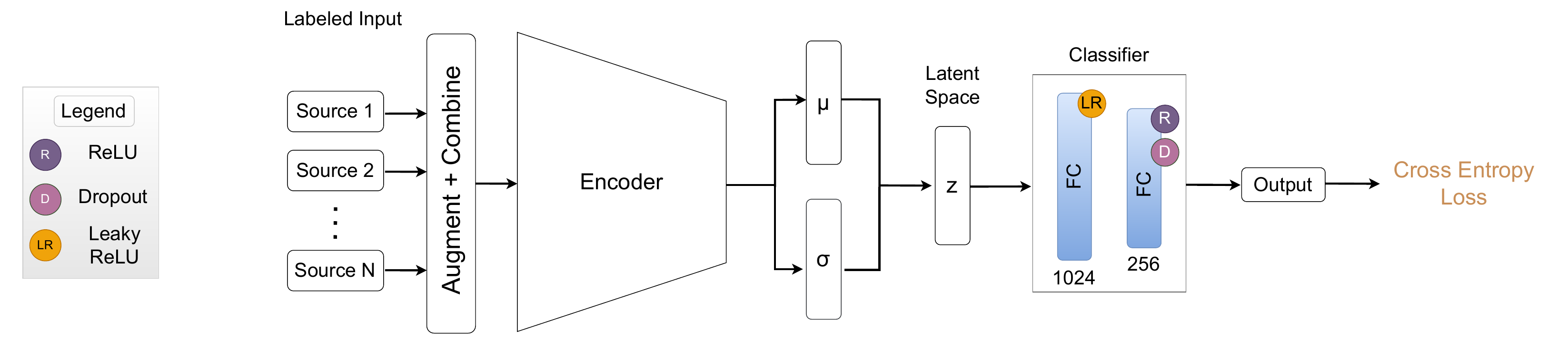}
\vspace{0.5em}
\caption{VaCDA classifier pipeline; The labeled input from source domains is processed by the pre-trained encoder and optimized by cross-entropy loss.}
\label{fig:arc2}
\end{figure*}

\section{Related Works}
\label{sec:related}

Domain adaptation (DA) is a well-established topic in the field of HAR \cite{Cook2013TransferLF}. Several approaches have successfully applied DA techniques to address distributional differences in activity recognition tasks. However, MSDA is less common than single-source DA. In this section, we review some of the related works.

\subsection{Discrepancy-based DA Approaches}
Data distribution heterogeneity among the data sources is measured by employing the divergence and distance-based metrics on the extracted features such as Maximum Mean discrepancy (MMD), KL divergence, and JSD-divergence. \cite{long2015learningtransferablefeaturesdeep} proposes DAN, which minimizes the domain discrepancy by embedding features into a reproducing kernel Hilbert space (RKHS) and performing multi-kernel MMD to reduce the difference between the domains.~\cite{10.1145/3360774.3360831} presents AugToAct, a semi-supervised approach combined with data augmentation \cite{10.1145/3360774.3360831,10.1145/3136755.3136817} techniques, uses affine transformations and Gaussian noise to enhance feature robustness.~\cite{sun2016deepcoralcorrelationalignment} present Deep CORAL, which implements DA by aligning the second-order statistics of source and target domains through correlation loss \cite{sun2016coral}. However, none of these methods specifically handle multi-source adaptation.


\subsection{Adversarial Learning-based DA Approaches}
There are several works that utilize adversarial learning to address DA~\cite{wilson2020multisourcedeepdomainadaptation,CHAKMA2021100174}.~\cite{wilson2020multisourcedeepdomainadaptation} investigates a multi-source approach, CoDATS, that leverages weak supervision across multiple source domains to improve the adaptation to a target domain.~\cite{FARIDEE2022100226} presents STranGAN, which uses a Spatial Transformer Network (STN) \cite{jaderberg2016spatialtransformernetworks} to perform adversarially-learned linear transformations, aligning the target with the source domain. The classifier training in StranGAN is separate from the spatial transformer training, whereas CoDATS trains its classifier within a single unified framework.

\subsection{VAE-based DA Approaches}
Only a limited number of studies on domain adaptation in HAR have leveraged the generalization capability~\cite{pmlr-v139-havtorn21a} of VAE.~\cite{9475454} develops AdaptNet, which tackles domain adaptation in HAR by using a semi-supervised deep translation network. This network aligns feature spaces of source and target domains through VAEs and Generative Adversarial Networks (GANs). Another VAE-based approach by~\cite{8949726} address the challenge of user-specific variability by using active learning to choose uncertain samples for labeling, which helps reduce the amount of labeled data needed from new users. However, this work does not particularly focus on domain adaptation.

\subsection{Contrastive Learning-based DA Approaches}
Contrastive learning is a versatile method for domain adaptation, typically applied in a self-supervised manner without labeled data, but it can also be extended to supervised settings to improve feature discrimination across domains. \cite{wilson2023caldaimprovingmultisourcetime} introduces CALDA, a multi-source domain adaptation (MSDA) model that combines contrastive and adversarial learning to align features across source domains and an unlabeled target domain, effectively narrowing the domain gap. CALDA achieves state-of-the-art performance in diverse distribution settings. More recently, CrossHAR \cite{hong2024crosshar} tackles distribution heterogeneity in cross-dataset HAR by using a hierarchical self-supervised pretraining framework, incorporating data augmentation, masked sensor modeling, and contrastive regularization to learn generalizable representations.




\section{Methodology}
\label{sec:vacda}
The VaCDA architecture consists of two pipelines: 1) a pre-training pipeline that aligns features across all domains, and 2) a classifier pipeline that classifies activities based on these aligned features, as shown in Figures \ref{fig:arc1} and \ref{fig:arc2}. The pre-training pipeline combines a VAE and a projection head for contrastive learning. By utilizing all available domains, the model learns to generalize the feature space effectively. The encoder minimizes the Kullback-Leibler (KL) divergence to capture features across distributions, while the projection head ensures discriminative features for different activities. Notably, the encoder is not updated during classifier training. VaCDA is a scalable solution, capable of handling any number of source domains. The following sections describe the problem formulation, our framework, and the adaptation mechanism.

\subsection{Problem Setup}

We define multi-source domain adaptation (MSDA) between heterogeneous feature spaces as follows. Let $(D_{s_j})_{j=1}^{S}$ represent $S$ source domains, where each source domain $D_{s_j} = {(x_{s_j}^{(i)}, y_{s_j}^{(i)})}_{i=1}^{n_{s_j}}$ consists of labeled samples $x_{s_j}^{(i)} \in {\rm I\!R}^{d}$, with corresponding labels $y_{s_j}^{(i)} \in C$ and $n_{s_j}$ samples. The target domain $D_t = {(x_t^{(i)}, y_t^{(i)})}_{i=1}^{n_t}$ contains $n_t$ unlabeled samples $x_t^{(i)} \in {\rm I\!R}^{d}$. While all domains share the same class set $C$ and dimensionality $d$, their feature distributions may differ from one another. Specifically, the distributions of the source domains $P_j(X_{s_j})$ for $j=1, \dots, S$ may not align with each other, nor with the distribution of the target domain $P(X_t)$, creating challenges for knowledge transfer. The goal of MSDA is to classify the target domain samples $x_t^{(i)}$ using the labeled source domains $(D_{s_j})_{j=1}^{S}$ despite these distributional differences.

\subsection{Proposed Framework}\label{proposed_framework}
The overall VaCDA architecture is shown in Figures \ref{fig:arc1} and \ref{fig:arc2}. In the following sections, we discuss the core principles and major components of our architecture.

(1) \textbf{Variational Autoencoder (VAE)}: 
With the VAE, we aim to align the source domains $(D_{s_j})_{j=1}^{S}$ and the target $D_t$ domain with a common distribution $D = (x^{(i)})_{i=1}^{n}$, where each $x^{(i)} \in {\rm I\!R}^{d}$.  We assume that each $x^{(i)}$ is sampled from a generative process $p(x|\boldsymbol{z})$, where $\boldsymbol{z}$ represents latent variables. In practice, latent variables $\boldsymbol{z}$ and the generative process $p(x|\boldsymbol{z})$ are unknown. The goal of the variational autoencoder is to jointly train two components: an inference network $q_\phi(\boldsymbol{z}|x)$ and a generator network $p_\theta(x|\boldsymbol{z})$. In the VAE framework, the inference network is implemented as a neural network that outputs the parameters for a Gaussian distribution, $q_\phi(\boldsymbol{z}|x)=\mathcal{N}(\mu_\phi(x),\Sigma_\phi(x))$. The generator network, $f_\theta(z)$, is a deterministic neural network parameterized by $\theta$, and the generative density $p_\theta(x|\boldsymbol{z})$ is modeled as a Gaussian distribution: $p_\theta(x|\boldsymbol{z})=\mathcal{N}(f_\theta(\boldsymbol{z}),\sigma^2I$). The VAE is trained by minimizing the negative Evidence Lower Bound (ELBO), defined as follows:

\begin{align} \label{eq:1}
    \mathcal{L}_{\text{ELBO}}(\theta, \phi; x) = - \mathbb{E}_{z \sim q_\phi(\boldsymbol{z}|x)}[\log p_\theta(x|\boldsymbol{z})] \nonumber \\
    - KL[q_\phi(\boldsymbol{z}|x)||p(\boldsymbol{z})]
\end{align}

Here, $p(\boldsymbol{z})$ represents the prior, which is modeled as a standard normal distribution $\mathcal{N}(0, I)$. The first term, $\mathbb{E}_{z \sim q_\phi(\boldsymbol{z}|x)}[\log p_\theta(x|\boldsymbol{z})]$ is simply the reconstruction loss between the input signal and the generated signal by the generator network $f_\theta(\boldsymbol{z})$. The second term, KL divergence acts as a regularizer, ensuring that the learned distribution $q_\phi(\boldsymbol{z}|x)$ stays close to the prior $\mathcal{N}(0, I)$. Both networks are optimized together by minimizing the ELBO:
\begin{equation} \label{eq:2}
\mathcal{L}_{VAE} = \min_{\theta, \phi} \, \mathbb{E}_{x \sim p_{\text{data}}(x)} \, \mathcal{L}_{\text{ELBO}}(\theta, \phi; x)
\end{equation}
Here, $p_{\text{data}}$ represents the distribution defined by the training set $T = D_{s_1} \cup D_{s_1} \cup \dots \cup D_{s_S} \cup D_t$.

(2) \textbf{Contrastive Learning}: Our framework uses two different contrastive losses i) \textit{self-supervised contrastive loss} (inspired by SimCLR framework), and ii) \textit{supervised contrastive loss}. The self-supervised contrastive loss helps the model learn domain-invariant features by contrasting positive pairs (augmentations of the same data) with negative pairs (dissimilar instances). we can describe it mathematically as follows. Let $\boldsymbol{x_i}$ and $\boldsymbol{x_j}$ be two augmented views of the same input $\boldsymbol{x}$ ($\boldsymbol{x} \sim p_{\text{data}}$), and $\boldsymbol{x_k}$ be a different instance. The latent representations $\boldsymbol{z_i}$, $\boldsymbol{z_j}$, and $\boldsymbol{z_k}$ are drawn from the latent space through the probabilistic encoder $q_\phi(\boldsymbol{z}|x)$. Specifically, $\boldsymbol{z_i} \sim q_\phi(\boldsymbol{z}|x_i)$, $\boldsymbol{z_j} \sim q_\phi(\boldsymbol{z}|x_j)$, and $\boldsymbol{z_k} \sim q_\phi(\boldsymbol{z}|x_k)$.
Additionally, the projected representations $\boldsymbol{r_i}$, $\boldsymbol{r_j}$, and $\boldsymbol{r_k}$ are the transformations of $\boldsymbol{z_i}$, $\boldsymbol{z_j}$, and $\boldsymbol{z_k}$, respectively, obtained through a projection head. The self-supervised loss:
\begin{equation} \label{eq:3}
    \mathcal{L}_{con}^{self} = - \sum{i \in \boldsymbol{I}} \log \frac{\exp(\text{sim}(r_i, r_j)/\tau)}{\sum_{k \in I} \exp(\text{sim}(r_i, r_k)/\tau)}
\end{equation}
Where $\boldsymbol{I}$ is the set of all instances in the batch. $sim(r_i, r_j)$ represents the cosine similarity between $\boldsymbol{r_i}$ and $\boldsymbol{r_j}$, which are the projections of the latent representations. $\tau$ is a temperature parameter controlling the sharpness of the distribution.

Additionally, we propose employing a supervised contrastive function to learn the domain-specific features within each source domain \cite{khosla2021supervisedcontrastivelearning}. We define positive pairs as samples with the same label across the source domain(s) and negative pairs as samples with different labels. The contrastive loss is applied as follows:
\begin{multline} \label{eq:4}
    \mathcal{L}_{con}^{sup} = -\frac{1}{N} \sum_{i=1}^{N} \left[ \log \left( \frac{\sum_{j=1}^{N} \exp\left(sim(r_i,r_j)\right) \cdot L_{ij}}{\sum_{j=1}^{N} \exp\left(sim(r_i,r_j)\right)} \right) \right. \\
    + \left. \log \left( \frac{\sum_{j=1}^{N} \exp\left(sim(r_i,r_j)\right) \cdot (1 - L_{ij})}{\sum_{j=1}^{N} \exp\left(sim(r_i,r_j)\right)} \right) \right]
\end{multline}
Here, $N$ is the batch size. The first log term encourages vectors with the same label \((L_{ij} = 1)\) to have higher similarity, as it rewards high values for the numerator (positive pair similarities) and minimizes the loss when the positive pairs are dominant. The second log term discourages high similarity between negative pairs \((L_{ij} = 0)\), as it minimizes the loss when the similarities of negative pairs are low, pushing them apart in the latent space. $\boldsymbol{sim_{i,j}}$ is the similarity between the projected representations $\boldsymbol{r_i}$ and $\boldsymbol{r_j}$, computed as a similarity matrix (e.g., cosine similarity) between all instances $\boldsymbol{i}$ and $\boldsymbol{j}$ in the batch. eq. Equations \ref{eq:3} and \ref{eq:4} complement each other, with eq. \ref{eq:3} capturing domain-invariant features across all domains, and eq. \ref{eq:4} focusing on domain-specific features within the source domains.

(3) \textbf{Classifier}: The classifier is a softmax-based neural network used for multi-class classification. It operates on latent representations generated by a previously trained classifier. During training, both the original and augmented representations are passed through the classifier, and cross-entropy loss is computed for both to update the model.
\begin{equation}\label{eq:5}
    \mathcal{L}_{clf} = \frac{1}{|b|} \sum_{i=1}^{|b|} \left( CE(\hat{y}_i, y_i) + CE(\hat{y}_i^{aug}, y_i) \right)
\end{equation}
Where, $\hat{y} = \boldsymbol{C}(z_i), \hat{y}_i^{aug} = \boldsymbol{C}(z_i^{aug})$, $y_i$ is the true label, and $CE$ is the cross-entropy loss.

\subsection{Optimization and Label Prediction}
We define our optimization function using equations \ref{eq:2}, \ref{eq:3}, and \ref{eq:4} to train the encoder and the decoder of the contrastive VAE as follows:
\begin{equation}\label{eq:6}
    \mathcal{L} = \mathcal{L}_{VAE} + \mathcal{L}_{con}^{self} + \mathcal{L}_{con}^{sup}
\end{equation}

\renewcommand{\algorithmicrequire}{\textbf{Input:}}
\renewcommand{\algorithmicensure}{\textbf{Output:}}
\begin{algorithm}[tb]
\caption{VAE training with Contrastive Learning}\label{alg:1}
\begin{algorithmic}[1]
    \REQUIRE minibatch of $m$ samples containing source and target domain data.
    \ENSURE A well-trained VAE that establishes a shared latent space between the source and target domains.
    \STATE Initialize the VAE: encoder$= \boldsymbol{E}$, decoder$= \boldsymbol{D}$, and projector$=\boldsymbol{P}$
        \FOR{each epoch}
            \FOR{each batch in total\_batches}
                \STATE sample a batch $x$ consisting of $b$-sized instances from $x_s, x_t$, where, $(x_s,y_s) \in (D_{s_j})_{j=1}^{S}$ and $x_t \in D_t$
                \STATE create the augmented version of $x$, $x^{aug}$
                \STATE pass the samples through the encoder: $\boldsymbol{E}(x), \boldsymbol{E}(x^{aug})$
                \STATE sample latent vectors by reparameterization trick: $z, z^{aug}$
                \STATE reconstruct samples: $D(z), D(z^{aug})$
                \STATE Calculate VAE loss eq:(\ref{eq:2}): $\mathcal{L}_{VAE}$
                \STATE Project latent vectors: $r = P(z)$ and $r^{aug} = P(z^{aug})$
                \STATE calculate self-supervised contrastive loss eq.\ref{eq:3}: $\mathcal{L}_{con}^{self}$
                \STATE calculate supervised contrastive loss with the source domain data eq.\ref{eq:4}: $\mathcal{L}_{con}^{sup}$
                \STATE update encoder and decoder with eq:(\ref{eq:6}): $\mathcal{L} = \mathcal{L}_{VAE} + \mathcal{L}_{con}^{self} + \mathcal{L}_{con}^{sup}$
            \ENDFOR
        \ENDFOR
    \RETURN encoder, $\boldsymbol{E}$
\end{algorithmic}
\end{algorithm}
The pretraining process is shown in Algorithms \ref{alg:1}. It includes training the VAE with contrastive learning. For the classifier training,  we pass source domains through the pretrained encoder, $\boldsymbol{E}$, to obtain their latent vectors, which are then used to train the classifier with cross-entropy loss. For prediction. we simply pass the target domain to the same encoder + classifier pipeline, Fig. \ref{fig:arc2}.

\section{Experiments}
\label{sec:experiments}
In this section, we describe the datasets, and provide details of our experiments.

\subsection{Datasets}
We utilize four publicly available datasets: 1) DSADS~\cite{misc_daily_and_sports_activities_256}, 2) OPPORTUNITY~\cite{misc_opportunity_activity_recognition_226}, 3) PAMAP2~\cite{misc_pamap2_physical_activity_monitoring_231}, and 4) WISDM~\cite{misc_wisdm_smartphone_and_smartwatch_activity_and_biometrics_dataset__507}. 

\begin{enumerate}
    \item \textbf{DSADS:} We select 10 unique activities, excluding similar activities from the full activity set. The activities from 5 different body positions (Torso, RA,
LA, RL, LL) are considered and the activities are standing, lying on the back, ascending, walking in the parking lot, treadmill running, stepper exercise, cross-trainer exercise, rowing, jumping, and playing basketball.
    \item \textbf{OPPORTUNITY:} We select all the available activities from 4 different users (each user has 5 different body positional - BACK, RUA, RLA, LUA, LLA data).
    \item  \textbf{PAMAP2:} Selected activities: lying, sitting, standing, walking, vacuuming, and ironing. These six activities are the most common across users. User-3 is excluded for missing activities.
    \item \textbf{WISDM:} For the WISDM (Smartphone and Smartwatch Activity) dataset, we select 8 activities---walking, jogging, stairs, sitting, standing, kicking, catching, and dribbling from subjects that have balanced datasets. The activity selection is based on a visual inspection of the correlation matrix, focusing on activities with lower feature correlations. Although no explicit cut-off is used, the observed patterns align with research indicating that high feature correlations can hinder domain adaptation \cite{JMLR:v17:15-239}. This is evident from the significant performance drop across all models when using the full set of 18 activities, highlighting the challenges posed by highly correlated features.

\end{enumerate}

\subsection{Baselines}
To evaluate VaCDA's domain adaptation performance, we compare it against six baselines: (i) STranGAN \cite{FARIDEE2022100226}, (ii) CoDATS \cite{wilson2020multisourcedeepdomainadaptation}, (iii) Deep CORAL \cite{sun2016deepcoralcorrelationalignment}, (iv) DAN \cite{long2015learningtransferablefeaturesdeep}, (v) CALDA \cite{wilson2023calda}, and (vi) CrossHAR \cite{hong2024crosshar}. All baselines are deep learning-based, as traditional machine learning models fall short in domain adaptation tasks. We focus exclusively on unsupervised learning approaches for a fair comparison. CoDATS and CALDA are multi-source adversarial methods, while Deep CORAL and DAN are well-known discrepancy-based techniques. CALDA represents the state-of-the-art (S.O.T.A.) for domain adaptation in human activity recognition (HAR). CrossHAR, a recent contrastive learning approach using transformers \cite{vaswani2017attention}, is included for its novel use of cross-dataset adaptation. STranGAN, a generative model with divergence-based techniques, also serves as a baseline. We adapt all models using their public repositories: CoDATS is converted from TensorFlow \cite{abadi2016tensorflow} to PyTorch \cite{paszke2017automatic}, Deep CORAL and DAN are adapted with custom feature extractors for IMU sensor data without changing their core concepts. CrossHAR, designed for cross-dataset heterogeneity (e.g., DSADS to PAMAP2), is easily extended to multi-source setups, as datasets like DSADS include multiple domains. For other single-source methods, we merge the domains.


\subsection{Evaluation Criteria}
Since accuracy, micro precision, micro recall, and micro F1 score yield similar results, we choose the micro F1 score for our experiments. This metric balances precision and recall, making it ideal for imbalanced datasets. \textit{To ensure reliable results, each experiment is run with three random seeds.}

\begin{table*}[h!]
    \centering
    \begin{tabular}{ p{1.5cm}  p{3.5cm} c c c c c c c c}
      \toprule
      Dataset & Task (Source $\rightarrow$ Target) & STranGAN & CoDATS & CALDA & D.CORAL & DAN & CrossHAR & VaCDA\\ 
      \midrule
      \multirow{5}{*}{DSADS} & RA, LA, RL, LL$\rightarrow$Torso & 44.08 & 47.85 & 48.95 & 39.67 & 63.02 & 41.78 & \textbf{66.94} \\
      & Torso, LA, RL, LL$\rightarrow$RA & 60.67 & 68.17 & 66.01 & 60.65 & 66.32 & 46.72 & \textbf{73.02} \\
      & Torso, RA, RL, LL$\rightarrow$LA & 59.87 & 64.95 & 65.94 & 54.93 & \textbf{78.18} & 57.26 & 75.85 \\
      & Torso, RA, LA, LL$\rightarrow$RL & 51.05 & 73.33 & \textbf{84.14} & 43.22 & 62.19 & 58.08 & 63.07 \\
      & Torso, RA, LA, RL$\rightarrow$LL & 53.83 & 63.19 & 65.06 & 41.97 & 59.09 & 60.54 & \textbf{66.77} \\
      \midrule
      DSADS & \textbf{Average} & 53.9 & 63.50 & 66.02 & 48.09 & 65.76 & 52.88 & \textbf{69.13}\\
      \bottomrule
      \\[1pt]
      \multirow{10}{*}{OPPO.} & RUA, RLA, LUA, LLA $\rightarrow$ BACK & 56.6 & 56.31 & 60.8 & 52.52 & 56.14 & 60.8 & \textbf{61.45} \\
      & BACK, RLA, LUA, LLA $\rightarrow$ RUA & 58.8 & 64.93 & \textbf{69.41} & 65.50 & 57.04 & 63.26 & 67.09 \\
      & BACK, RUA, LUA, LLA $\rightarrow$ RLA & 61.92 & 79.20 & \textbf{79.73} & 66.28 & 68.54 & 54.07 & 61.36 \\
      & BACK, RUA, RLA, LLA $\rightarrow$ LUA & 60.75 & 69.78 & 68.35 & 62.34 & 68.42 & 62.04 & \textbf{71.73} \\
      & BACK, RUA, RLA, LUA $\rightarrow$ LLA & 57.45 & 79.87 & \textbf{81.86} & 65.19 & 71.66 & 57.00 & 74.96 \\
      \midrule
      OPPO. & \textbf{Average} & 59.10 & 70.02 & \textbf{72.03} & 62.37 & 64.36 & 59.43 & 67.32\\
      \bottomrule
      \\[1pt]
      \multirow{3}{*}{PAMAP2} & Chest, Ankle $\rightarrow$ Wrist & 26.52 & 20.89 & 20.1 & 26.38 & 26.52 & 35.95 & \textbf{50.53} \\
      & Wrist, Ankle $\rightarrow$ Chest & 25.61 & 30.32 & 31.37 & 36.24 & 37.7 & 33.28 & \textbf{53.23} \\
      & Wrist, Chest $\rightarrow$ Ankle & 19.08 & 22.76 & 25.13 & 15.84 & 24.79 & 28.09 & \textbf{33.48} \\
      \midrule
      PAMAP2 & \textbf{Average} & 23.74 & 24.66 & 25.53 & 26.15 & 29.67 & 32.44 & \textbf{45.75}\\
      \bottomrule
    \end{tabular}
    \caption{Cross-Position Heterogeneity (Average of 3 runs per experiment; Metric: Micro F1 score).}
    \label{table:2}
\end{table*}


\begin{table*}[h!]
    \centering
    \begin{tabular}{ p{1.5cm}  p{3.5cm} c c c c c c c}
      \toprule
      Dataset & Task (Source $\rightarrow$ Target) & STranGAN & CoDATS & 
      CALDA & D.CORAL & DAN & CrossHAR & VaCDA\\ 
      \midrule
      \multirow{3}{*}{DSADS} & 1, 2, 3, 4, 5, 6, 7 $\rightarrow$ 8 & 66.93 & 82.69 & \textbf{85.28} & 55.4 & 77.26 & 77.28 & 78.16 \\
      & 1, 2, 3, 4, 5, 6, 8 $\rightarrow$ 7 & 60.80 & \textbf{78.47} & 75.71 & 55.19 & 74.85 & 70.88 & 69.41 \\
      & 1, 2, 3, 4, 5, 7, 8 $\rightarrow$ 6 & 60.97 & 80.93 & \textbf{86.96} & 47.93 & 83.96 & 69.89 & 71.79 \\
      \midrule
      DSADS & \textbf{Average} & 62.90 & 80.70 & \textbf{82.65} & 52.84 & 78.69 & 73.95 & 73.12\\
      \bottomrule
      \\[1pt]
      \multirow{3}{*}{OPPO.} & 1, 2, 3 $\rightarrow$ 4 & 78.05 & 81.87 & \textbf{84.40} & 72.84 & 79.18 & 57.40 & 74.38 \\
      & 1, 2, 4 $\rightarrow$ 3 & 72.99 & 77.18 & \textbf{77.78} & 61.42 & 68.91 & 58.47 & 70.96 \\
      & 1, 3, 4 $\rightarrow$ 2 & 76.79 & \textbf{79.12} & 79.01 & 70.32 & \textbf{75.72} & 55.13 & 71.65 \\
      \midrule
      OPPO. & \textbf{Average} & 75.94 & 79.39 & \textbf{80.39} & 68.19 & 74.60 & 57.00 & 72.33\\
      \bottomrule
      \\[1pt]
      \multirow{3}{*}{WISDM} & 1, 5, 7, 8, 12, 13, 15$\rightarrow$37 & 43.04 & 45.38 & 45.31 & 31.25 & 50.16 & 33.68 & \textbf{54.60} \\
      & 1, 5, 7, 8, 12, 13, 15$\rightarrow$38 & 30.42 & 52.61 & 53.38 & 24.90 & 46.32 & 34.00 & \textbf{59.91} \\
      & 1, 5, 7, 8, 12, 13, 15$\rightarrow$39 & 42.30 & 54.69 & \textbf{60.27} & 37.24 & 47.91 & 42.00 & 54.66 \\
      \midrule
      WISDM & \textbf{Average} & 38.59 & 50.89 & 52.98 & 31.13 & 48.13 & 36.56 & \textbf{56.39}\\
      \bottomrule
    \end{tabular}
    \caption{Cross-Person Heterogeneity (Average of 3 runs per experiment; Metric: Micro F1 score).}
    \label{table:3}
\end{table*}

\begin{table*}[h!]
    \centering
    \begin{tabular}{ p{1.5cm}  p{3.5cm} c c c c c c c }
      \toprule
      Dataset & Task (Source $\rightarrow$ Target) & STranGAN & CoDATS & CALDA & D.CORAL & DAN & CrossHAR & VaCDA\\ 
      \midrule
      \multirow{2}{*}{WISDM} & Phone $\rightarrow$ Watch & 40.02 & 48.37 & 49.27 & 45.97 & 53.1 & 43.04 & \textbf{55.74} \\
      & Watch $\rightarrow$ Phone & 34.89 & 40.58 & 40.45 & 40.77 & 42.85 & 33.47 & \textbf{43.60} \\
      \midrule
      WISDM & \textbf{Average} & 37.46 & 44.48 & 44.86 & 43.37 & 47.98 & 38.26 & \textbf{49.67} \\
      \bottomrule
    \end{tabular}
    \caption{Cross-Device Heterogeneity (Average of 3 runs per experiment; Metric: Micro F1 score).}
    \label{table:4}
\end{table*}

\subsection{Implementation Details}
    We implement all our models with PyTorch \cite{10.5555/1953048.2078195}. Experiments are run on a Linux Ubuntu 22.04.4 LTS system with CUDA 11.8, featuring a Ryzen 5800H CPU, 16GB RAM, and an Nvidia RTX 3060 GPU with 6GB VRAM. In preprocessing, we extract and standardize accelerometer data by body position, remove NaN entries, and split the data into training, validation, and test sets with a 60-20-20\% ratio to ensure balanced activity representation. A windowing technique with a 100-size window and 10\% overlap is applied. For data augmentation, we use: (i) jittering, (ii) scaling, (iii) time warping, and (iv) rotation. We use the ADAM optimizer \cite{kingma2017adammethodstochasticoptimization} with the following hyperparameters: 20 epochs, a learning rate of 0.0001 for the encoder and decoder, and 0.001 for the rest. The learning rate is constant for the first 10 epochs and reduced by a factor of 0.1 every 5 epochs. We create a batch from $S$ source domains and one target domain, shuffling data for balanced sampling before training. A batch size of 32 is used per domain, with a reconstruction loss weight of $10^6$ and a temperature value of 0.5 for contrastive losses.

\subsection{Experimental Setup} To evaluate the VaCDA framework, we consider three scenarios involving data distribution heterogeneity: (1) cross-position heterogeneity, (2) cross-person heterogeneity, and (3) cross-device heterogeneity. Below, we describe our experimental scenarios.
\begin{itemize}
    \item \textbf{Cross-position Heterogeneity:} This setup evaluates VaCDA using data from all users across multiple positions (e.g., wrist, hand, leg), with one position as the target domain and others labeled. DSADS, OPPORTUNITY, and PAMAP2 are used for evaluation.
    \item \textbf{Cross-person Heterogeneity:} To test VaCDA’s adaptability to new individuals, we use labeled data from multiple users across all body positions. This setup helps the model generalize to different movement patterns, ensuring strong performance on unseen individuals. DSADS, OPPORTUNITY, and WISDM are used for evaluation.
    \item \textbf{Cross-device Heterogeneity:} We evaluate VaCDA for domain adaptation between a smartphone and a smartwatch using data from all selected users. This experiment is conducted on the WISDM dataset as it is the only one that includes data from multiple devices.
\end{itemize}
\textit{For the sake of reproducibility, the code will be made publicly available upon acceptance of the paper.}

\section{Results \& Discussion}
\label{sec:result&analysis}
\subsection{Results}
\label{sec:results}
    VaCDA beats the baseline in cross-position setup (Table \ref{table:2}) on DSADS and PAMAP2 datasets achieving 69.13\% and 45.75\% respectively with a notable 13\% boost on PAMAP2. However, all models perform poorly on PAMAP2. This issue may stem from data misalignment across body positions, as noted in \cite{CHAKMA2021100174}, possibly due to activity labeling or sensor measurement issues. Due to that reason, we chose not to include PAMAP2 in the cross-position setting. In terms of performance, all models perform better on cross-person (Table \ref{table:3}) setup compared to cross-position (Table \ref{table:2}) and cross-device (Table \ref{table:4}). While it falls short on DSADS and OPPORTUNITY (cross-person), VaCDA shows a 6\% improvement over baselines on WISDM. In cross-device setting VaCDA marginally outperforms the baselines. Among the baselines, CALDA stands out, particularly in the cross-person setup, followed by the CoDATS. CrossHAR, being the most recent work among the baselines, its performance on multi-source domain adaptation is unexpectedly average and it is beaten by VaCDA in almost all scenarios. Experiments show that DAN and Deep CORAL occasionally outperform newer models, particularly STranGAN, which underperforms in most setups. Perhaps it is partly due to the custom feature extractor, as they are not originally designed for IMU.

\noindent
\textbf{Key Takeaways}
\begin{enumerate}
    \item VaCDA outperforms the baselines in cross-position setup by 3\%-13\% (DSADS and PAMAP2).
    \item VaCDA demonstrates competitive performance in cross-position setup (outperforms baselines on WISDM).
    \item VaCDA marginally beats the baselines in cross-device setting (WISDM).
\end{enumerate}

\subsection{Ablation Study}
\begin{figure}[!ht]
    \centering
    \subfloat[Cross-Position Heterogeneity]{\includegraphics[width=.23\textwidth]{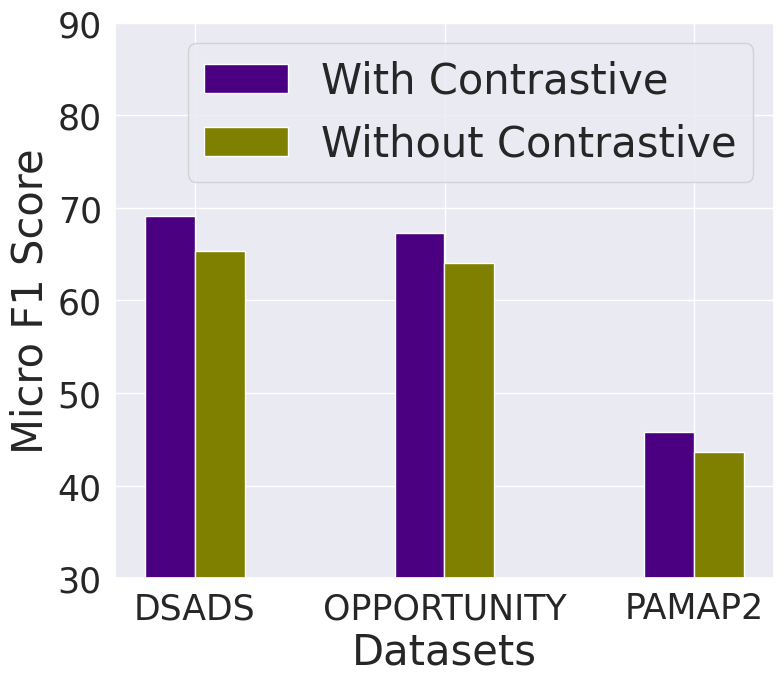}}\quad
    \subfloat[Cross-Person Heterogeneity]{\includegraphics[width=.23\textwidth]{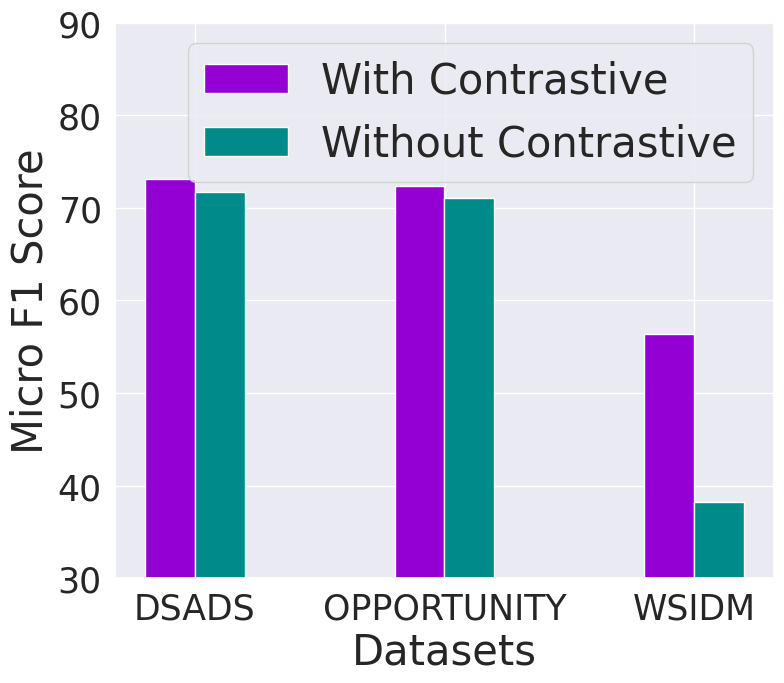}}
    \caption{VaCDA performance comparison with and without contrastive learning}
    \label{fig:sub1}
\end{figure}

\label{sec:contrastive_analysis}
We examine how contrastive learning influences domain adaptation performance. To run these experiments, we remove both contrastive functions from our optimization equation \ref{eq:6}. Figure \ref{fig:sub1} highlights the performance gains achieved by incorporating contrastive losses into the VaCDA framework. The graph on the left, Figure \ref{fig:sub1}(a) demonstrates that contrastive learning consistently enhances the micro F1 score across all datasets for cross-position heterogeneity, with improvements ranging from 2.08\% to 3.75\%. On the right, Figure \ref{fig:sub1}(b) shows a similar comparison for cross-person heterogeneity, where contrastive learning boosts F1 scores across all datasets: DSADS (+1.43\%), OPPORTUNITY (+1.32\%), and WISDM (+18.19\%). WISDM sees the most substantial improvement. This study highlights the value of integrating contrastive learning into the VaCDA framework.




\section{Conclusion}
\label{sec:conclusion}
We introduce a novel MSDA framework, VaCDA, which combines variational autoencoders with contrastive learning to tackle data distribution discrepancies under challenging and real-life scenarios in wearable human activity recognition. The framework effectively mitigates domain discrepancies by utilizing the generalization and discriminative capabilities of its components, facilitating better adaptation to diverse data distributions. Simultaneously, it solves the scalability issue of handling multiple source domains. VaCDA shows notable improvements in cross-position and cross-device heterogeneity. While it does not outperform the baselines in the cross-person scenario, it still achieves competitive performance. In the future, we will focus on extending the framework to address complex research challenges, such as open-set domain adaptation, and generalized novel category recognition.




\bibliographystyle{IEEEtran}

\end{document}